\documentclass{bmvc2k}

\title{A Robust Normalizing Flow using Bernstein-type Polynomials}

\usepackage{cleveref}
\usepackage{enumitem}
\usepackage{comment}
\setlist[itemize]{topsep=-3pt}

\usepackage{thmtools}

\usepackage{graphicx, wrapfig} 
\usepackage{booktabs} 

\usepackage{amssymb,amsmath,amsthm}

\newtheorem{theorem}{Theorem}

\usepackage{mathtools} 
\newtheorem{exmp}{Example}

\newtheorem{definition}{Definition}
\newtheorem{remark}{Remark}
\usepackage{nicefrac}  
\addauthor{Sameera Ramasinghe$^*$}{sameera.ramasinghe@anu.edu.au}{1}
\addauthor{Kasun Fernando$^*$}{kasun.akurugodage@utoronto.ca}{2}
\addauthor{Salman Khan}{salman.khan@mbzuai.ac.ae}{3}
\addauthor{Nick Barnes}{nick.barnes@anu.edu.au}{1}

\addinstitution{
Australian National University
}
\addinstitution{
University of Toronto
}
\addinstitution{
 Mohamed bin Zayed University of AI
}

\runninghead{Ramasinghe \textit{et al.}}{A Robust NF using Bernstein-type Polynomials}

\usepackage{cleveref}

\usepackage{enumitem}
\usepackage{comment}
\setlist[itemize]{topsep=-3pt}

\usepackage{thmtools}

\usepackage{graphicx, wrapfig} 
\usepackage{booktabs} 

\usepackage{amssymb,amsmath,amsthm}

\begin{document}

\maketitle

\begin{abstract}
Modeling real-world distributions can often be challenging due to sample data that are subjected to perturbations, \textit{e.g.}, instrumentation errors, or added random noise. Since flow models are typically nonlinear algorithms, they amplify these initial errors, leading to poor generalizations. This paper proposes a framework to construct Normalizing Flows (NFs) which demonstrate 
higher robustness against such initial errors. To this end, we utilize Bernstein-type polynomials inspired by the optimal stability of the Bernstein basis. Further, compared to the existing NF frameworks, our method provides compelling advantages like theoretical upper bounds for the approximation error, better suitability for compactly supported densities, and the ability to employ higher degree polynomials without training instability.  We conduct a  theoretical analysis and empirically demonstrate the efficacy of the proposed technique using experiments on both real-world and synthetic datasets.\looseness=-1
\end{abstract}

\vspace{-0.5em}
\section{Introduction}
\vspace{-0.5em}
Modeling the probability distribution of a set of observations, \emph{i.e.}, generative modeling, is a crucial task in machine learning. It enables the generation of synthetic samples using the learned model and allows the estimation of the likelihood of a data sample. 
This field has met with great success in many problem domains including image generation~\citep{ho2019flow++, kingma2018glow, lu2020woodbury}, audio synthesis~\citep{esling2019universal, prenger2019waveglow}, reinforcement learning~\citep{mazoure2020leveraging, ward2019improving}, noise modeling~\citep{abdelhamed2019noise}, and simulating physics experiments~\citep{wirnsberger2020targeted, wong2020gravitational}. In the recent past, deep neural networks such as generative adversarial networks (GANs) and variational autoencoders (VAEs) have been widely adopted in generative modeling due to their 
success in modeling high dimensional distributions. However, they entail several limitations:~1) exact density estimation of arbitrary data points is not possible, and 2) training can be cumbersome due to aspects such as mode collapse, posterior collapse and high sensitivity to architectural design of the model~\citep{Kobyzev_2020}.  

In contrast, normalizing flows (NFs) are a category of generative models that enable exact density computation and efficient sampling (for theoretical foundations see Appendix 1.1 and references therein). As a result, NFs have been gaining increasing attention from the machine learning community since the seminal work of \citet{rezende2015variational}. 
In essence, NFs consist of a series diffeomorphisms that 
transforms a simple distribution into a more complex one, 
and must be designed so that  the Jacobian determinants of these diffeomorphisms can be efficiently calculated (This is, in fact, an essential part of the implementation). To this end, two popular approaches have been proposed so far: 1) efficient determinant calculation methods such as \citet{berg2018sylvester, grathwohl2018ffjord, lu2020woodbury}, and 2) \emph{triangular maps} 
~\citep{jaini2019sum, dinh2014nice, dinh2016density}. 
The key benefit of triangular maps is that their Jacobian matrices are triangular, and hence, the calculation of Jacobian determinants takes only $O(n)$ steps as opposed to the $O(n^3)$ complexity of the computation of a determinant of an unconstrained $n\times n-$matrix. In this paper, we focus only on triangular maps. 

On the one hand, it is not a priori clear whether such a constrained class of maps is expressive enough to produce sufficiently complex transformations. Interestingly, ~\citep{bogachev2005triangular} showed that, given two probability densities, there exists a unique 
increasing triangular map that transforms one density 
to the other. Consequently, the constructed NFs should be \textit{universal}, \textit{i.e.}, dense in the class of increasing triangular maps, in order to approximate those density transformations with arbitrary precision. 
But it is observed in \citet{jaini2019sum} that, despite many NFs being triangular, they are not universal. 
To remedy this, most
have reverted to the empirical approach 
of stacking several transformations together, 
thereby increasing the expressiveness of the model. 
Alternatively, there are NFs that use 
genuinely universal transformations. 
Many such methods employ 
coupling functions based on polynomials, 
\textit{e.g.}, sum-of-squares (SOS) polynomials in \citet{jaini2019sum}, cubic splines in \citet{durkan2019cubic} or rational quadratic splines in \citet{durkan2019neural}. 
Here, we 
employ another class of polynomials called Bernstein-type polynomials to construct a universal triangular flow which henceforth is called Bernstein-type NF. Our universality proof is a consequence of \cite{Bernstein}, and unlike the proofs in the previous literature, is constructive, and hence, 
yields analytic expressions for approximations of known density transformations; see \Cref{sec:universality}.

On the other hand, noise is omnipresent in data. Sample data can be subjected to perturbations due to experimental uncertainty (instrumentation errors or added random noise). It is well-known that nonlinear systems 
amplify these initial errors and produce drastically different outcomes even for small changes in the input data; see \citet{Errors, Numerics, Classification}. In terms of (deep) classifiers, robustness is often studied in the context of adversarial attacks, where the performance of the classifier should be robust against specific perturbations of the inputs. Similarly, for generative models, this translates to robustly modeling a distribution in the presence of perturbed data. Recently, \cite{condessa2020provably} analyzed the robustness of deep generative models against random perturbations of the inputs, where they designed a VAE variant that is robust to random perturbations. Similarly, \cite{kos2018adversarial} also proposed a heuristic-based method to make deep generative models robust against perturbations of the inputs.\looseness=-1

As any other nonlinear model, NFs are also susceptible of numerical instabilities.  
Unless robust, trained NFs 
may amplify 
initial errors, and demonstrate out-of-distribution sample generation and poor generalization to unseen data. For instance, consider an NF modeling measured or generated velocities (energies) of molecular movements; see \cite{kohler2020equivariant}.
Similarly, consider a scenario where we intend to flag certain samples of a random process as out-of-distribution data. If the training data is susceptible to noise, the measured log-likelihoods of the test samples may significantly deviate from the true values unless the NF model is robust.


Therefore, it is imperative that the robustness of NFs is investigated during their construction and implementation. Despite this obvious importance, robustness of NFs has not been theoretically or even experimentally studied in the previous literature, unlike other deep generative models. One of the key motivations behind this work is to fill this void. 
Accordingly, we show that Bernstein-type polynomials are ideal candidates for the construction of NFs that are not only universal 
but also robust. Robustness of Bernstein-type NFs follow from the \textit{optimal stability} of the Bernstein basis ~\citep{Farouki2, Farouki1}; see also, \Cref{sec:robust}. 

Recently, Bernstein polynomials have also been used in conditional transformation models to  due to their versatility; see for example, \citet{Horthorn2018}, \citet{Horthorn2021}, \cite{Baumann2021} and references therein. In contrast, here, we introduce a novel approach of building NFs using Bernstein polynomials. \looseness=-1

In summary, apart from collecting, organizing and summarising in a coherent fashion 
the appropriate theoretical results which were scattered around the mathematical literature, we \textbf{1)} deduce, in \Cref{Universality}, the universality of Bernstein flows, \textbf{2)} state and prove, in \Cref{the:monotonic}, a \textit{strict} monotonicity result of Bernstein polynomials which has been mentioned without proof and used in \citet{Farouki2000}, \textbf{3)} prove, in \Cref{thm:cpttarget}, that, in \textit{any} NF, it is enough to consider compactly supported targets (in the previous literature was implicitly assumed without proper justification), \textbf{4)} theoretically establish that, compared to other polynomial-based flow models, Bernstein-type NFs 
    demonstrate superior robustness to perturbations in data. To our knowledge, ours is the first work to discuss robustness in NFs, \textbf{5)} discuss a theoretical bound for the rate of convergence of Bernstein-type NFs, which, to our knowledge, has not been discussed before in the context of NFs, and \textbf{6)} propose a practical framework to construct normalizing flows using Bernstein-type polynomials and empirically demonstrate that theoretically discussed properties hold in practice.


Moreover, compared to previous NF models, our method has several additional advantages such as suitability for approximating compactly supported target densities; see \Cref{sec:range}, the ability to increase the expressiveness by increasing the polynomial degree at no cost to the training stability; see \Cref{sec:range}, and being able to invert easily and accurately due to the availability of efficient root finding algorithms; see \Cref{sec:monotonicity}. \vspace{-10pt}

\section{Theoretical foundations of the Bernstein-type NF}
\label{sec:theory}
Here, we elaborate on the desirable properties of Berstein-type polynomials and their implication to our NF model. The mathematical results taken directly from existing literature are stated as facts with appropriate references. The proofs of Theorems \ref{thm:cpttarget}, \ref{the:monotonic} and \ref{Universality}, appear in Appendix 2. 
Also, in each subsection, we point out the advantages of our model (over existing models) based on the properties discussed. We point the reader to Appendix 1.1 
for a brief discussion on triangular maps and other preliminaries.\vspace{-5pt} 
\subsection{Bernstein-type polynomials}\label{sec:bernstein}
 The degree $n$ Bernstein polynomials, 
    $\binom{n}{k} x^k(1-x)^{n-k},\, 0 \leq k \leq n,\, n \in \mathbb{N}$,
were first introduced by Bernstein in his constructive proof of the Weierstrass theorem in \citet{Bernstein}. 
In fact, given a continuous function $f:[0,1]\to \mathbb{R}$, its degree $n$ Bernstein approximation, $B_n(f):[0,1] \to \mathbb{R}$, given by
\vspace{-5pt} 
\begin{equation}\label{BernAppr}
B_n(f)(x)=\sum_{k=0}^n f\left(\frac{k}{n}\right) \binom{n}{k} x^{k}(1-x)^{n-k},\vspace{-5pt}
\end{equation}
is such that $B_n(f)\to f$ uniformly in $[0,1]$ as $n \to \infty$. Moreover, Bernstein polynomials form a basis for degree $n$ polynomials on $[0,1]$.
More generally, polynomials of Bernstein-type can be defined as follows.\vspace{-5pt} 
\begin{definition}
A degree $n$ polynomial of Bernstein-type is a polynomial of the form
\vspace{-5pt} 
\begin{equation}\label{BernPoly}
    B_n(x)=\sum_{k=0}^n \alpha_k \binom{n}{k} x^{k}(1-x)^{n-k},\,\,x \in [0,1]\,,\vspace{-5pt} 
\end{equation}
where $\alpha_k$, $0\leq k \leq n$ are some real constants.
\end{definition}
\begin{remark}\label{rem:GeneralBern}
Polynomials of Bernstein-type on an arbitrary closed interval $[a,b]$ are defined by composing $B_n$ with the linear map that sends $[a,b]$ to $[0,1]$, $L_{a,b}(x)=\frac{x-a}{b-a}.$
So, Bernstein-type polynomials on $[a,b]$ take the form $B_n \circ L_{a,b}$.
Hereafter, we denote degree $n$ Bernstein-type polynomials by $B_n$ regardless of the domain.\vspace{-5pt} 
\end{remark}
As we shall see below, one can control various properties of Bernstein-type polynomials like strict monotonicity, range and universality by specifying conditions on the coefficients, and the error of approximation depends on the degree of the polynomials used.\vspace{-5pt} 

\subsection{
Easier control of the range and suitability for compact targets}\label{sec:range}

The supports of distributions of samples used when training and applying NFs are not fixed. So, it is important to be able to easily control the range of the coupling functions. In the case of Bernstein-type polynomials, $B_n$s, this is very straightforward. Note that if $B_n$ is defined on $[a,b]$, then $B_n(a)=\alpha_0$ and $B_n(b)=\alpha_n$. Therefore, one can fix the values of a Bernstein-type polynomial at the end points of $[a,b]$ by fixing $\alpha_0$ and $\alpha_n$. So, if $B_n$ is increasing (which will be the case in our model; see \Cref{sec:monotonicity}), then its range is $[\alpha_0,\alpha_n
]$. This translates to a significant advantage when training for compactly supported targets because we can achieve any desired range $[c,d]$ (the support of the target) by fixing $\alpha_0=c$ and $\alpha_n=d$ and letting only $\alpha_k,\, 0<k<n$ vary. So, $B_n$s are ideal for modeling compactly supported targets. In fact, in most other methods except splines in \citet{durkan2019cubic, durkan2019neural}, either there is no obvious way to control the range or the range is infinite. 
We present the following theorem to establish that for the purpose of training, we can assume the target has compact support (up to a known diffeomorphism). \looseness=-1
\vspace{-0.5em}
\begin{theorem}\label{thm:cpttarget} Let $I_j, j=1,2,3$ be measurable subsets of $\mathbb{R}^d$. Suppose $I_1$ is the support of the target $P_x$, $I_2$ is the support of the prior $P_z$, $\mathfrak{F}$ is the class of coupling functions with ranges contained in $I_3$ and $h:I_3 \to I_1$ is a diffeomorphism. If $P_y$ is the distribution on $I_3$ such that $h_*P_y = P_x$, then
\vspace{-5pt} 
\begin{align}
   \text{\emph{arg min}}_{f \in \mathfrak{F}}\, \text{\emph{KL}}(P_x \| (h \circ f)_*P_z ) = \text{\emph{arg min}}_{f \in \mathfrak{F}}\, \text{\emph{KL}}(P_{y} \| f_*P_z ).
\end{align}
\end{theorem}
In the previous literature that uses transformations with compact range, this fact was implicitly assumed without proper justification. As a consequence of the above theorem, our coupling functions having finite ranges is not a restriction, and in \textit{any} NF model, even if the target density is not compactly supported, the learning procedure can be implemented by first converting the target density to a density with a suitable compact support via a diffeomorphism, and then training on the transformed data. Since we deal with compactly supported targets, in practice, we do not need to construct deep architectures (with a higher number of layers), as we can increase the degree of the polynomials to get a better approximation. In other polynomial based methods, a practical problem arises because the higher order polynomials could predict extremely high values initially leading to unstable gradients (\textit{e.g.}, \cite{jaini2019sum}). In contrast, we can avoid that problem as the range of our transformations can be explicitly controlled from the beginning by fixing $\alpha_0$ and $\alpha_n$ . 

\subsection{Strict monotonicity and efficient inversion} 
\label{sec:monotonicity}

In triangular flows, the coupling maps are expected to be invertible. 
Since strict monotonocity implies invertibility, it is sufficient that the $B_n$s we use are strictly monotone. 
\vspace{-0.5em}
\begin{theorem}
\label{the:monotonic}
Consider the Bernstein-type polynomial $B_n$ in \eqref{BernPoly}. Suppose $\alpha_0 < \alpha_1 < \dots < \alpha_n$. Then, $B_n$ is strictly increasing on $[0,1]$. \vspace{-5pt} 
\end{theorem}

This result was mentioned as folklore without proof and used in \citet{Farouki2000}. On the other hand, in \citet{Coupling}, the conclusion is stated with monotonicity and not \textit{strict} monotonicity (which is absolutely necessary for invertibility). Hence, we added a complete proof of the statement in Appendix 2. 
According to this result, the strict monotonicity of $B_n$s depends entirely on the strict monotonicity of the coefficients $\alpha_k$s. It is easy to see that the assumption of strict monotonicity of the coefficients is not a further restriction on the optimization problem. 
For example, if the required range is $[c,d]$, we can take $\alpha_{n-k}=c+(d-c)(1+v^2_0+\dots+v^2_k)^{-1}$
where $v_k$s are real valued. This converts the constrained problem of finding $\alpha_k$s to an unconstrained one of finding $v_k$s.  Alternatively, we can take $\alpha_0=c$ and $\alpha_k =|v_1|+\dots + |v_k|$, and after each iteration, linearly scale $\alpha_k$s in such a way that $\alpha_n=d$.
After guaranteeing invertibility, we focus on computing the inverse, 
\textit{i.e.}, at each iteration, given $x$ we solve for $z \in [0,1]$,
\vspace{-5pt}
\begin{equation}
   B_n(z)= \sum_{k=0}^n \alpha_k \binom{n}{k} z^{k}(1-z)^{n-k} = x  \iff  \sum_{k=0}^n (\alpha_k-x) \binom{n}{k} z^{k}(1-z)^{n-k} = 0\vspace{-5pt} 
\end{equation}
because Bernstein polynomials form a partition of unity on $[0,1]$. So, finding inverse images, \emph{i.e.}, solving 
the former is equivalent to finding solutions to the latter. 
Due to our assumption of increasing $\alpha_k$s, $B_n$ is increasing, and has at most 
one root on $[0,1]$. The condition $(\alpha_0-x)(\alpha_n-x)<0$ (which can be easily checked) guarantees the existence of a unique solution, 
and hence, the invertibility of the original transformation. 

Due to the extensive use of Bernstein-type polynomials in computer-aided geometric design, there are several well-established efficient root finding algorithms at our disposal ~\citep{PhDSpencer}. For example, the parabolic hull approximation method in \citet{RajanReport} is ideal for higher degree polynomials with fewer roots (in our case, just one) and has cubic convergence for simple roots (better than both the bijection method and Newton's method). Further, because of the numerical stability described in \Cref{sec:robust} below, the use of Bernstein-type polynomials in our model minimizes the errors in such root solvers based on floating–point arithmetic. Even though inverting splines are easier due to the availability of analytic expressions for roots, compared to all other other NF models, we have more efficient and more numerically stable algorithms that allow us to reduce the cost of numerical inversion in our setting.\looseness=-1\vspace{-5pt}


\subsection{Universality and the explicit rate of convergence} \label{sec:universality}

In order to guarantee universality of triangular flows, we need to use a class of coupling functions that well-approximates increasing continuous functions. 
This is, in fact,  the case for $B_n$s, and hence, we have the following theorem whose proof we postpone to Appendix 2.\vspace{-5pt}  
\vspace{-1em}
\begin{theorem}\label{Universality}
Bernstein-type normalizing flows are universal.
\end{theorem} 
\vspace{-0.5em}

The basis of all the universality proofs of NFs in the existing literature is that the learnable class of functions is dense in the class of increasing continuous functions. In contrast, the argument we present here is constructive. As a result, we can write down sequences of approximations for (known) transformations between densities explicitly; see Appendix 4.

In the case of cubic-spline NFs of \citet{durkan2019cubic}, it is known that for $k=1,2,3$ and $4$, when the transformation is $k$ times continuously differentiable and the bin size is $h$, the error is $O(h^k)$ \citep[Chapter 2]{SplinesBook}. However, we are not aware of any other instance where an error bound is available. 
Fortunately for us, the error of approximation of a function $f$ by its Bernstein polynomials has been extensively studied. 
We recall from \citet{Voronovskaya} the following error bound: for $f:[0,1] \to \mathbb{R}$ twice continuously differentiable
\vspace{-5pt}
\begin{equation}
\scriptsize
    B_n(f)-f=\frac{x(1-x)}{2n}f''(x)+o(n^{-1})\vspace{-5pt}
\end{equation}
and this holds for an arbitrary interval $[a,b]$ with $x(1-x)$ replaced by $(x-a)(b-x)$.
Since the error estimate is given in terms of the degree of the polynomials used,
we can improve the optimality of our NF by avoiding unnecessarily high degree polynomials. This allows us to keep the number of trainable parameters under control in our NF model.
It can be shown that the error $O(n^{-1})$ 
above does not necessarily improve when SOS polynomials are used instead; see Appendix 3. 
In our NF, at each step, the estimation is done using a univariate polynomial, and hence, the overall convergence rate is, in fact, the minimal univariate convergence rate of $O(n^{-1})$ (equivalently, the error upper bound is the maximum of univariate upper bounds),
and in general, cannot be improved further regardless of how regular the density transformation is. However, our experiments (in \Cref{sec:numericalerror}) 
show that our model on average has a significantly smaller error than the given theoretical upper-bound.\vspace{-5pt} 

\subsection{Robustness of Bernstein-type normalizing flows}\label{sec:robust}


In this section, we recall some known results in \citet{Farouki2, Farouki1} about the optimal stability of the Bernstein basis. The two key ideas are that smaller \emph{condition numbers} lead to smaller numerical errors and that the Bernstein basis has the
optimal condition numbers compared to other polynomial bases.


To illustrate this, let $p(x)$ be a polynomial on $[a,b]$ of degree $n$ expressed in terms of a basis $\{\phi_k\}_{k=0}^n$, \textit{i.e.}, \vspace{-5pt}
\begin{equation}
    p(x)=\sum_{k=0}^n c_k \phi_k(x),\, x \in [a,b].\vspace{-5pt}
\end{equation}
Let $c_k$ be randomly perturbed, with perturbations $\delta_k$ where the relative error $\delta_k / c_k \in (-\varepsilon,\varepsilon)$. Then the total pointwise perturbation is\vspace{-5pt} 
\begin{equation}
    \delta(x) = \sum_{k=0}^n \delta_k \phi_k(x)
 \implies |\delta (x)| \leq \sum_{k=0}^n|\delta_k \phi_k(x)| \leq \varepsilon \sum_{k=0}^n|c_k\phi_k(x)|\leq \varepsilon C_\phi (p(x))\,,\vspace{-5pt}
\end{equation}
 where $C_\phi (p(x)) := \sum_{k=0}^n|c_k \phi_k(x)|$ is the condition number for the total perturbation with respect to the basis $\phi_k$. It is clear that $C_\phi (p(x))$ controls the magnitude of the total perturbation. 

According to \citet{Farouki2}, if $\phi=\{\phi_k\}_{k=0}^n$ and $\psi=\{\psi_k\}_{k=0}^n$ are non-negative bases for polynomials of degree $n$ on $[a,b]$, and for all $j$, latter is a non-negative linear combination of the former, that is, $\psi_j = \sum_{k=0}^n M_{jk}\phi_k$ with $M_{jk} \geq 0$. Then, for any polynomial $p(x)$, \vspace{-5pt}
\begin{equation}
    C_\phi(p(x))\leq C_\psi(p(x))\,.\vspace{-5pt}
\end{equation}
For $0\leq a < b$, the Bernstein polynomials and the power monomials, $\{1,x,x^2,\dots,x^n\}$, are non-negative bases on $[a,b]$. It is true that the latter is a positive linear combination of the former but \textit{not} vice-versa; see \citet{Farouki2}. 
Therefore, Bernstein polynomial basis has the lowest condition number out of the two.
This means that 
the change in the value of a polynomial caused by a perturbation of coefficients is always smaller in the Bernstein basis than in the power basis. A more involved computation gives \vspace{-5pt}
\begin{equation}
    \tilde{C}_\phi(x_0)=\left(\frac{m!}{|p^{(m)}(x_0)|}\sum_{k=0}^n|c_k\phi_k(x_0)|\right)^{1/m}\vspace{-5pt}
\end{equation}
as the condition number that controls the computational error for a $m-$fold root $x_0$ of $p(x)$ in $[a,b]$; see \citet{Farouki1}. There, it is proved that if  $\tilde{C}_\psi(x_0)$ and $\tilde{C}_{\phi}(x_0)$ denote the condition numbers for finding roots of \textit{any} polynomial on $[0,1]$ in the power and the Bernstein bases on $[0,1]$, respectively, then $\tilde{C}_\phi(x_0)<\tilde{C}_{\psi}(x_0)$ for $x_0\in(0,1]$ and $\tilde{C}_\phi(0)=\tilde{C}_{\psi}(0)$. 
This means that the change in the value of a root of a polynomial caused by a perturbation of coefficients is always smaller in the Bernstein basis than in the power basis.

In fact, a universal statement is true: Among all non-negative basis on a given interval, the Bernstein polynomial basis is \textit{optimally stable} in the sense that no other non-negative basis gives smaller condition numbers for the values of polynomials (see \cite[Theorem 2.3]{Pena1997}) and no other basis expressible as non-negative combinations of the Bernstein basis gives smaller condition numbers for the roots of polynomials (see \cite[Section 5.3]{Farouki2}).

In particular, $B_n$s are systematically more stable than the polynomials in the power form when determining roots (for example, when inverting) and evaluation (for example, when finding image points). As a result, when polynomials are used to construct NFs (for example, Q-NSF based on quadratic or cubic splines, SOS based on some of square polynomials and our NF based on Bernstein-type polynomials, ours yields the most numerically stable NF, i.e., it is theoretically impossible for them to be more robust. Our experiments in \Cref{sec:robustexp}, while confirming this, demonstrate that our method outperforms even the NFs that are \textit{not} based on polynomials.\looseness=-1\vspace{-5pt}

\vspace{-10pt} 
\section{Construction of the Bernstein-type normalizing flow}\label{sec:method}\vspace{-10pt}

\begin{figure}[!ht]
    \centering
    \includegraphics[width=0.7\columnwidth]{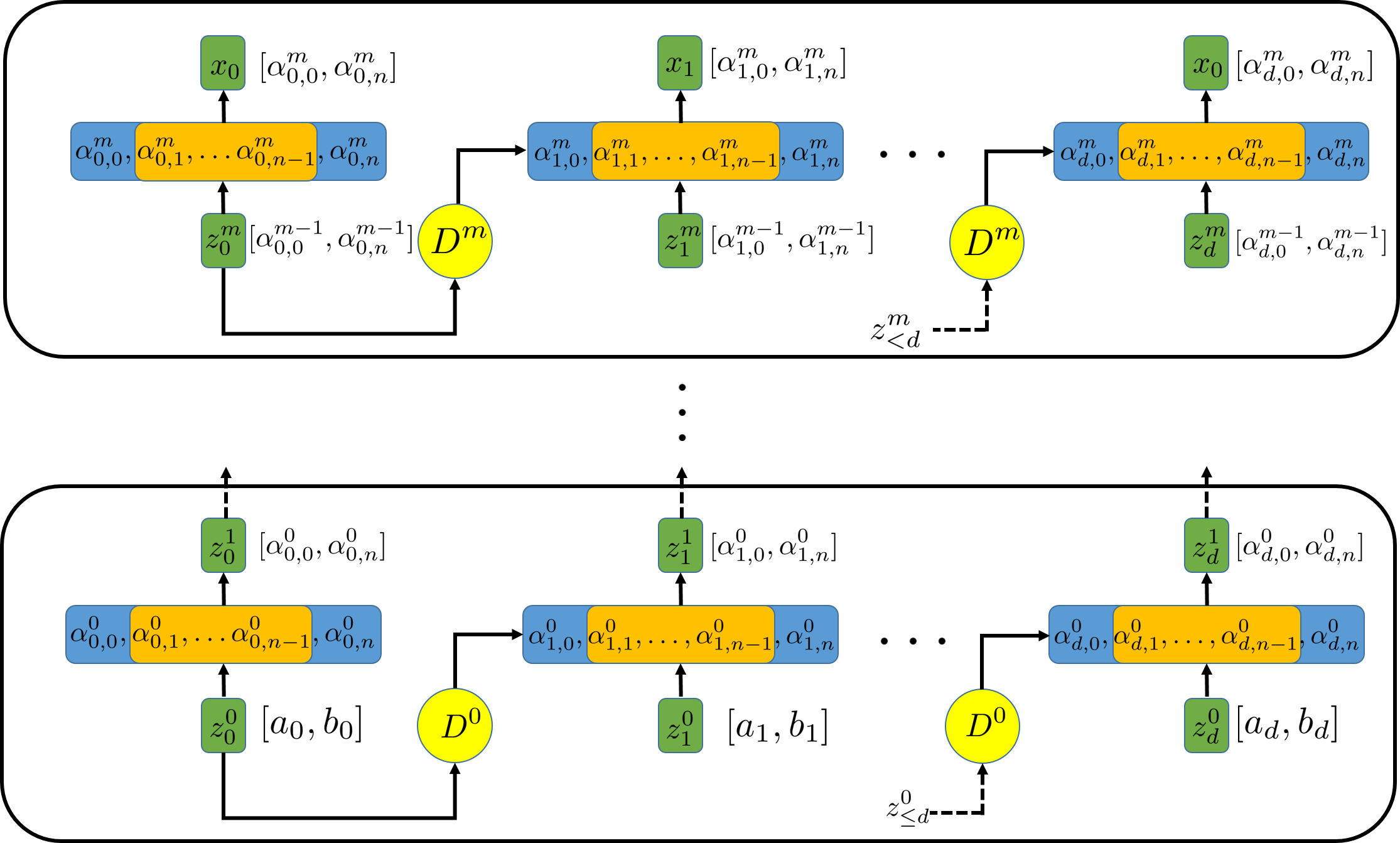}
    \vspace{-5pt} 
    \caption{\small Overall Bernstein NF architecture with $m+1$ layers for $d$-dimensional distributions. The ranges of transformations are within brackets and trainable coefficients are in orange boxes.}
    \label{fig:overall}
\end{figure}\vspace{-10pt} 

In this section, we describe the construction of normalizing flows using the theoretical framework established in \Cref{sec:theory} for compactly supported targets. For this, we employ a MADE style network \citep{germain2015made}. Consider a $d$-dimensional source $P_z(\textbf{z})$ and a $d$-dimensional target $P_x(\textbf{x})$. Then, the element-wise mapping between the components $x_j$ and $z_j$ is approximated using a Bernstein-type polynomial as $x_j = B^j_n(z_j)$. We obtain the parameters of $B^j_n(z_j)$ using a neural network which is conditioned on $z_{< j}$. This ensures a triangular mapping between the distributions. We fix $\alpha_0$ and $\alpha_n$ to be constants, and thus, define the range of each transformation; see \Cref{sec:range}. Moreover, as per \Cref{the:monotonic}, $\alpha_k$s need to be strictly increasing  for a  transformation to be strictly increasing. However, when we convert this constrained problem to an unconstrained one as proposed in \Cref{sec:monotonicity}, we obtain $v_k$s using the neural network and then calculate $\alpha_k$s as described. 

For each $B_n^j$, we employ a fully-connected neural net with three layers to obtain the parameters, except in the case of $B_n^0$ in which we directly optimize the parameters. Figure \ref{fig:overall} illustrates a model architecture with $m+1$ layers and degree $n$ polynomials for $d$-dimensional distributions. Here, there are $(n-1)(m+1)$ variable coefficients altogether. 
We use maximum likelihood to train the model with a learning rate $0.01$ with a decay factor of $10\%$ per $50$ iterations. All the weights are initialized randomly using a standard normal distribution. \vspace{-10pt}






\vspace{-5pt}
\section{Experiments}\label{sec:experiments}
\vspace{-5pt}
In this section, we summarise our empirical evaluations of the proposed model based on both real-world and synthetic datasets and compare our results with other NF methods.\vspace{-10pt}

\subsection{Modeling sample distributions} \vspace{-10pt}
\begin{table*}[ht]
\scriptsize
\caption{\small Test log-likelihood comparison against the state-of-the-art on real-world datasets (higher is better). Results for competing methods are extracted from \citet{durkan2019neural} where error bars correspond to
two standard deviations. Log-likelihoods are averaged over 5 trials for Bernstein.
}
\label{tab:real_world}
\vspace{-5pt}
\begin{center}
\begin{sc}
\begin{tabular}{lcccccr}
\toprule
Model  & Power & Gas & Hepmass & MiniBoone & BSDS300  \\
\midrule
FFJORD &  $0.46 \pm 0.01$  & $8.59 \pm 0.12$   & $-14.92 \pm 0.08$ & $-19.43 \pm 0.04$ & $157.40 \pm 0.19$  \\
GLOW & $0.42 \pm 0.01$ & $12.24 \pm 0.03$ & $-16.99 \pm 0.02$ & $-10.55 \pm 0.45$ & $156.95 \pm 0.28$\\
MAF & $0.45 \pm 0.01$ & $12.35 \pm 0.02$ & $-17.03 \pm 0.02$ & $-10.92 \pm 0.46$ & $156.95 \pm 0.28$     \\
NAF & $0.62 \pm 0.01$ & $11.96 \pm 0.33$ & $-15.09 \pm 0.04$ & $-8.86 \pm 0.15$ & $157.73 \pm 0.04$        \\
BLOCK-NAF & $0.61 \pm 0.01$ & $12.06 \pm 0.09$ & $-14.71 \pm 0.38$ & $-8.95 \pm 0.07$ & $157.36 \pm 0.03$ \\
RQ-NSF (AR) & $0.66 \pm 0.01$ & $13.09 \pm 0.02$ & $-14.01 \pm 0.03$ & $-9.22 \pm 0.48$ & $157.31 \pm 0.28$\\

Q-NSF (AR) & $0.66 \pm 0.01$ & $13.09 \pm 0.02$ & $-14.01 \pm 0.03$ & $-9.22 \pm 0.48$ & $157.31 \pm 0.28$\\
SOS & $0.60 \pm 0.01$ & $11.99 \pm 0.41$ & $-15.15 \pm 0.10$ & $-8.90 \pm 0.11$ & $157.48 \pm 0.41$ & \\
{BERNSTEIN} 
& $0.63 \pm 0.01$ & $12.81 \pm 0.01$ & $-15.11 \pm 0.02$ & $-8.93 \pm 0.08$ & $157.13 \pm 0.11$\\
\bottomrule
\end{tabular}
\end{sc}
\end{center}
\vspace{-15pt}
\end{table*}
\begin{wraptable}{r}{5.5cm}
\vspace{-10pt} 
\scriptsize
\caption{\small Test log-likelihood comparison against the state-of-the-art on image datasets (higher is better). Results for competing methods are extracted from \citet{jaini2019sum}. Note that the first three models use multi-scale convolutional architectures.}
\label{tab:images}
\vspace{-2pt} 
\begin{center}
\begin{sc}
\begin{tabular}{lcr}
\toprule
Model & MNIST & CIFAR10  \\
\midrule
Real-NVP & $-1.06$ & $-3.49$\\
FFJORD & $-0.99$ & $-3.40$\\
GLOW & $-1.05$ & $-3.35$\\
MAF & $-1.89$ & $-4.31$\\
MADE & $-2.04$ & $-5.67$\\
SOS & $-1.81$ & $-4.18$\\
{BERNSTEIN} & $-1.54$ & $-4.04$ \\
\bottomrule
\end{tabular}
\end{sc}
\end{center}\vspace{-15pt} 
\end{wraptable} 

We conducted experiments on four datasets from the UCI machine-learning repository and BSDS300 dataset. \Cref{tab:real_world} compares the obtained test log-likelihood against recent flow-based models. As illustrated, our model achieves competitive results on all of the five datasets. 
We observe that our model consistently reported a lower standard deviation which may be attributed to the robustness of our model.\looseness=-1

\begin{figure}
\centering
\begin{minipage}[b]{\dimexpr 0.5\textwidth-0.4\columnsep}
\centering
         \includegraphics[width=0.95\columnwidth]{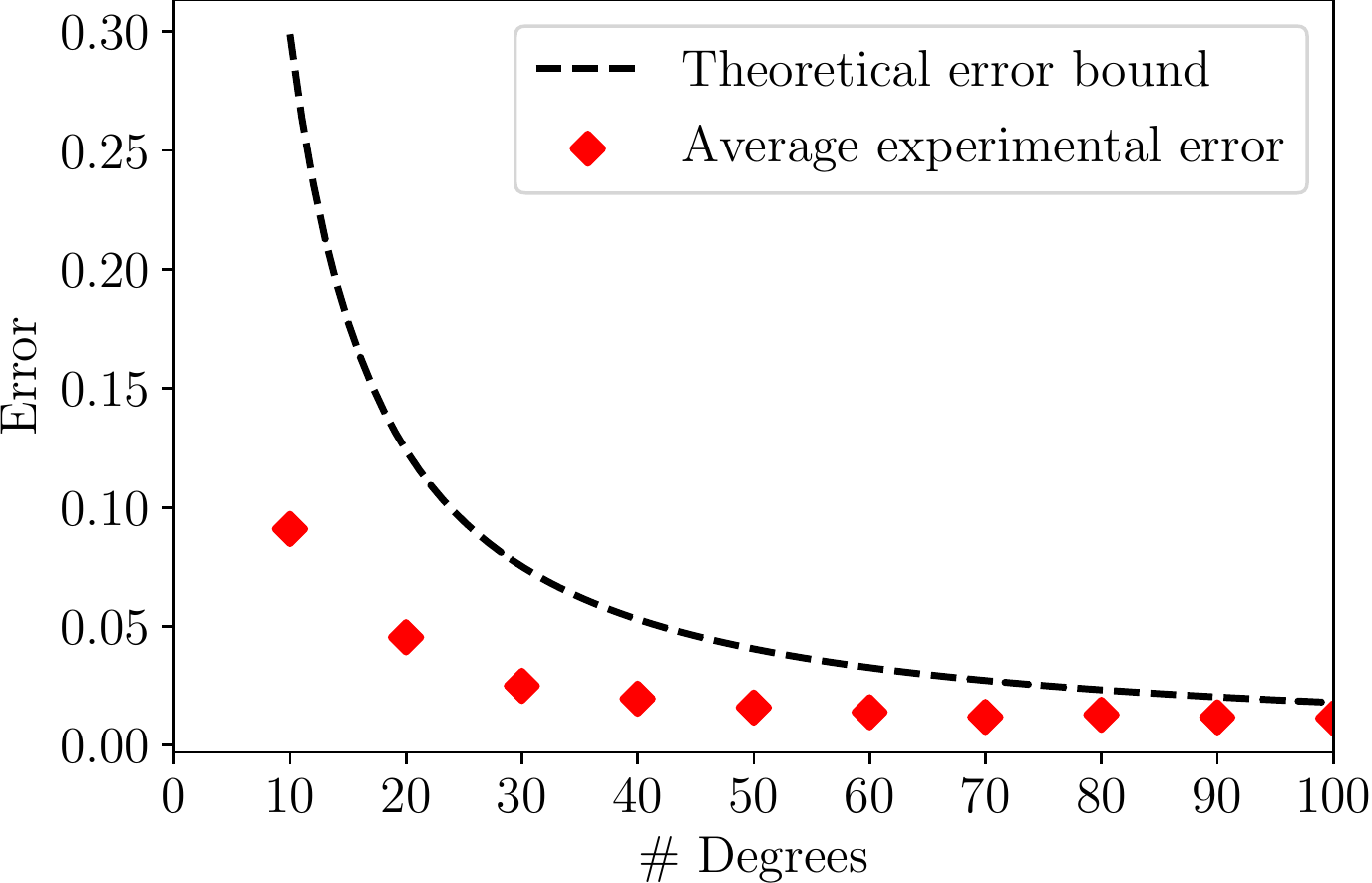}
   \vspace{-5pt} 
    \caption{\small 
    Error bound vs 
    average experimental error.}
    \label{fig:error_bound}
\end{minipage}%
 \hfill
\begin{minipage}[b]{\dimexpr 0.5\textwidth-0.4\columnsep}
\centering
         \includegraphics[width=0.85\columnwidth]{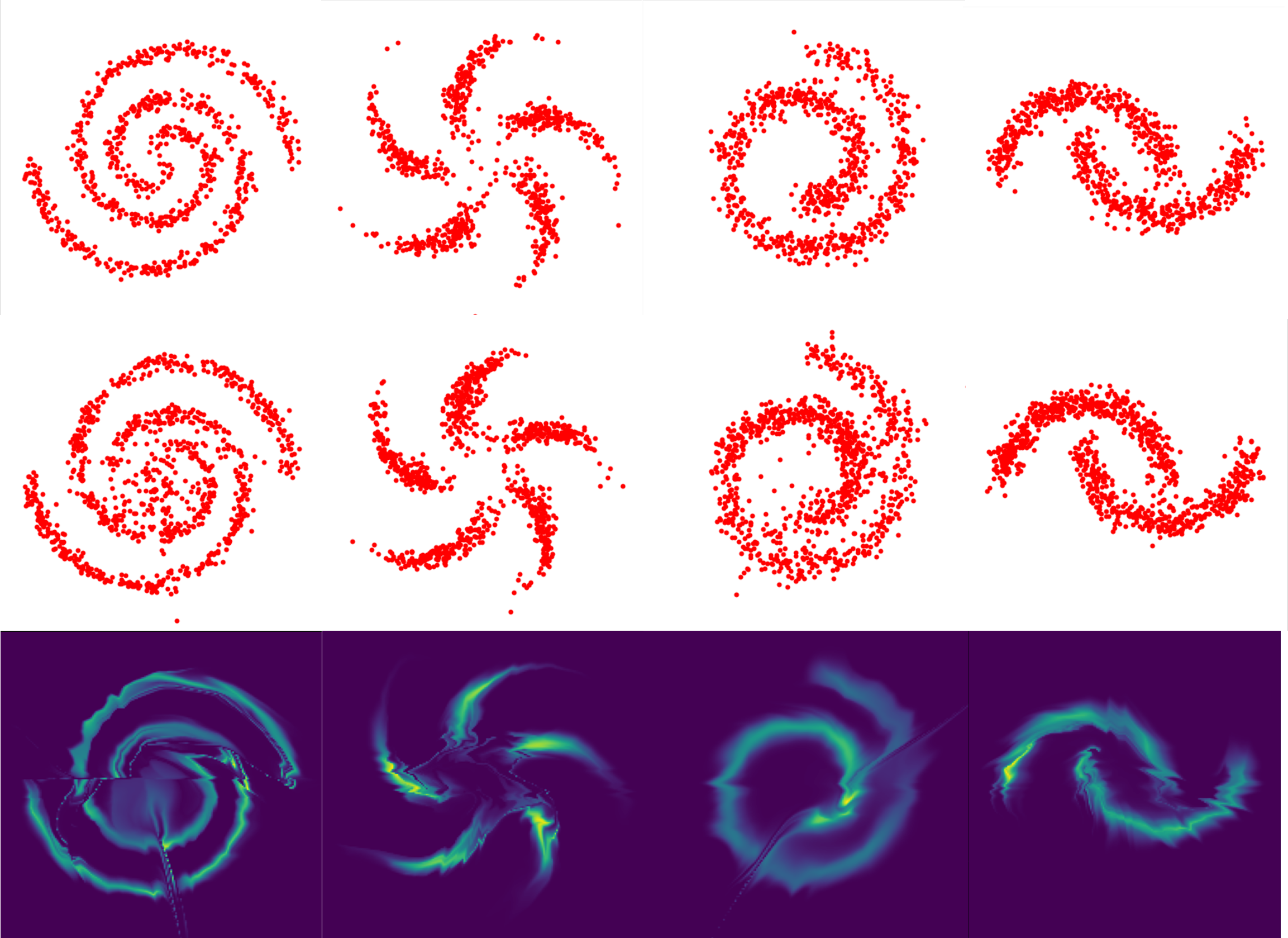}
   \caption{\small Qualitative results for modeling the toy distributions. \emph{From the top row}: ground truth, prediction, and predicted density.}
   \vspace{-5pt}
    \label{fig:toy_data}
\end{minipage}
\end{figure}

We also applied our method to two low-dimensional image datasets, CIFAR10 \& MNIST. The results are reported in \Cref{tab:images}. Among the methods that do not use multi-scale convolutional architectures, we obtain the optimal results. 
 In addition, we tested our model on several toy datasets (shown in \Cref{fig:toy_data}). Note that these 2D datasets contain multiple modes, sharp jumps and are not fully supported. So, the densities are not that obvious to learn. Despite the difficulties, our model is able to estimate the given distributions in a satisfactory manner.\looseness=-1

\vspace{-10pt}

\subsection{Robustness}\label{sec:robustexp}
In order to experimentally verify that Bernstein-type NFs are more numerically stable than other polynomial based NFs (as claimed in \Cref{sec:robust}) we use a standard idea in the literature; see \citet{Errors}. 

We add i.i.d.~noise, sampled from a Uniform$[0,10^{-2}]$, to the five datasets included in \Cref{tab:real_world}, and measure the change in the test log-likelihood as  a fraction of the standard deviation (so that the change is in terms of standard deviations). In practice, values of the experimented datasets are rescaled to a magnitude around unity. In signal processing, a good SNR is considered to be above 40DB which is used in most real-world cases. Here, we have chosen a noise order of $10^{-2}$ because our intention is to demonstrate that a SNR level below or even around that range can affect the performance of NFs.\looseness=-1

For a fair comparison, we train all the models from scratch on the noise-free train set using the codes provided by the authors, strictly following the instructions in the original works to the best of our ability. Then, we test the models on the noise-free test set. We run the above experiment $5$ times to obtain the standard deviation $\sigma$ and mean $\mu$ of the test log-likelihood. Next, we add noise to the training set, retrain the model, and obtain the test log-likelihood $y$ on the noise-free test set. Finally, we obtain the metric $\frac{|y-\mu|}{\sigma}$ which we report in Table 3.\vspace{-5pt}\looseness=-1

\begin{table}[h]\label{NoiseExp}
\scriptsize 
\caption{\small Test log-likelihood drop for random initial errors, relative to the original standard deviations.}
\vspace{-1em}
\label{tab:noisy}
\begin{center}
\begin{sc}
\begin{tabular}{lccccc}
\toprule
Model  & {Power} & {Gas} & {Hepmass} & {MiniBoone} & {BSDS300} \\
\midrule
{FFJORD} & $2.7$ & $4.4$ & $3.2$ & $1.7$ & $6.6$\\
{Real-NVP} & $2.4$ & $4.2$ & $3.6$ & $1.4$ & $7.4$\\
{GLOW} & $2.1$ & $4.1$ & $2.3$ & $0.8$ & $6.9$\\
NAF & $2.2$ & $3.7$ & $3.3$ & $0.7$ & $6.6$\\
{MAF}  & $2.4$ & $4.4$ & $3.9$ & $0.8$ & $7.1$\\
{MADE}  & $2.1$ & $4.6$ & $3.6$ & $2.4$ & $8.1$\\
{RQ-NSF}  & $2.3$ & $5.4$ & $4.1$ &  $0.9$ & $7.8$\\
{SOS} & $2.1$ & $1.7$ & $1.9$ & $1.6$ & $6.1$\\
{BERNSTEIN} & $1.1$ & $1.3$ & $1.1$ & $0.6$ & $2.3$\\
\bottomrule
\end{tabular}
\end{sc}
\end{center}
\vspace{-10pt}
\end{table}

As expected, Bernstein NF demonstrate the lowest relative change in performance, implying the robustness against random initial errors. In fact, other models are not robust: even small initial errors consistently (at least in 4 out of 5 datasets) created changes \textit{larger} than $1.645\,\sigma$ (corresponding to the two $5$\% tails of the distribution of errors) where $\sigma$ is the original standard deviation of error. In comparison, the change in our NF consistently (4 out of 5) was well-within the acceptable range and is at most $1.3\, \sigma$. In the remaining dataset, change in our model is $2.3\, \sigma$ while in all other models it's \textit{more} than $6\,\sigma.$\looseness=-1
\vspace{-10pt}
\subsection{Validation of the theoretical error upper-bound}\label{sec:numericalerror}
The degree $n\, (\geq 5)$ Bernstein approximation of $f \in C^3[0,1]$ has an error upper-bound \vspace{-5pt}
\begin{equation}E_n=n^{-1}\|\rho^2 f^{(2)}\|_\infty+n^{-3/2}\|\rho^3 f^{(3)}\|_\infty\vspace{-5pt}
\end{equation} where $\rho(x) = \sqrt{x(1-x)}$ \citep[Chapter 4]{BernOp}. 
Now, we verify this 
using a Kumarswamy$(2,5)$ distribution
as the prior 
and Uniform$[0,1]$ as the target. 
Let $f(x)=1-(1-x^2)^5$ and $B_n$ be the learned degree $n$ Bernstein-type polynomial. 
The average error, 
    $\int_0^1|f(x)-B_n(x)|\, dx,$
obtained using the learned $B_n$s and $E_n$ 
satisfying\vspace{-5pt} 
\begin{equation}
    1.25n^{-1} + 5n^{-3/2}<E_n < 1.25n^{-1} + 5.5n^{-3/2}\vspace{-5pt}
\end{equation}
are plotted in \Cref{fig:error_bound}. It shows that the observed (average) error is smaller than this theoretical upper-bound. 
In the NF, we have used a single layer and increased the degree of the polynomial from 10 to 100. The NF model was stable even when the degree 100 polynomial was used. So, this experiment also demonstrates that our model is, in fact, stable even when higher degree polynomials are used (as claimed in \Cref{sec:range}).\looseness=-1 \vspace{-10pt}



\section{Conclusion}\vspace{-5pt}

We propose a novel method to construct a universal autoregressive NFs with Bernstein-type polynomials as the coupling functions, and is the first instance of robustness of NFs being discussed. We show that Bernstein-type NFs posses advantages like true universality, robustness against initial and round-off errors, efficient inversion, having an explicit convergence rate, and better training stability for higher degree polynomials.


\section*{Appendix}
\section{Preliminaries}\label{sec:background}\vspace{-5pt}
In this section, we describe the general set up of triangular normalizing flow models.\vspace{-10pt}

\subsection{Normalizing flows and triangular maps}
NFs learn an invertible mapping between a prior and a more complex distribution (the target) in the same dimension. Typically, the prior is chosen to be a Gaussian with identity covariance or uniform on the unit cube, and the target is the one we intend to learn. Below, we present a summary of related ideas and refer the readers to \citet{jaini2019sum} and \citet{Kobyzev_2020} for a comprehensive discussion.

More formally, let $\textbf{z}$ and $\textbf{x}$ be sampled data from the prior with density $P_{z}$ and the target distribution with density $P_{x}$, respectively. Then, NFs learn a transformation $f$ such that $f(\textbf{z}) = \textbf{x}$ which is differentiable and invertible with a differentiable inverse. Such transformations are called diffeomorphisms and they allow the estimation of the probability density $P_x(\textbf{x})$ via the change of variables formula 
   $ P_{x}(\textbf{x})=P_{z}(f^{-1}\textbf{x})|\textbf{J}_{f}(f^{-1}\textbf{x})|^{-1}$
where $\textbf{J}_f$ is the Jacobian determinant of $f$. 

Given an independent and identically distributed (i.i.d.)~sample $\{x_1, \dots, x_n\}$ with law $P_x$, learning the target density $P_x$ and the transformation $f$ (within an expressive function class $\mathfrak{F}$) is done simultaneously via minimizing the Kullback-Leibler (KL) divergence between $P_x$ and the pushforward of $P_z$ under $f$ denoted by $f_*P_z$,\vspace{-5pt}
\begin{equation}
        \label{equ:KLdivergence}
        \min_{f \in \mathfrak{F}} \text{KL}(P_x \| f_*P_z ) = - \max_{f \in \mathfrak{F}} \int \log \frac{P_{z}(f^{-1}\textbf{x})}{|\textbf{J}_{f}(f^{-1}\textbf{x})|}\, \cdot \, P_x(\textbf{x})\, d\textbf{x}.\vspace{-5pt}
\end{equation}

Density estimation using \eqref{equ:KLdivergence} requires efficient calculation of the Jacobian as well as $f^{-1}$. Both can be achieved via constraining $f$ to be an \emph{increasing triangular map}. That is, taking $P_{x}(\textbf{x})$ to be a multivariate distribution where $\textbf{x} = (x_1, x_2, \dots, x_d)$, and the prior $P_z(\textbf{z})$ where $\textbf{z} = (z_1, z_2, \dots z_d)$, the components of $\textbf{x}$ are expressed as $x_j = f_j(z_1,z_2,\dots, z_j)$ for suitably defined transformations $f_j, j = 1,2, \dots, d$ where $f_j$ is increasing with respect to $z_j$. From now on, we denote $(z_1,z_2,\dots, z_j)$ by $z_{< j+1}$. In this case, the Jacobian determinant is the product $\prod_{j=1}^d\partial_{z_j} f_j$. Also, because $f_j$ is increasing in $z_j$, inversion can be done recursively starting from $f^{-1}_1$.\looseness=-1 \vspace{-10pt}


\section{Proofs}\label{app:proofs}\vspace{-5pt}
\begin{proof}[Proof of Theorem 1]
To illustrate the relevance of the theorem to our setting, we write down the details of the proof assuming that the learnable class $\mathfrak{F}$ is Bernstein-type polynomials. The same proof is true for any class of functions.

Take $I_j=[a_j,b_j]$ for $j=1,2,3$ with $a_j<b_j$ with the possibility that $a_1=-\infty$ or $b_1=\infty$ whence the interval is understood to be open on the infinite end and the target may have non-compact support. Let $B_n :I_2 \to I_3$ be the learnable Bernstein-type polynomial with coeffecients $\{\alpha_j\}_{j=0}^n$. Let $h:I_3\to I_1 $ be a fixed invertible transformation so that $h^{-1}$ transforms the target density $P_x$ to $P_y$ supported on $I_3$, i.e., $h_*P_y = P_x$, and\vspace{-5pt}
\begin{equation}
    P_x(\textbf{x}) = \frac{P_y(h^{-1}\textbf{x})}{|\textbf{J}_h(h^{-1}\textbf{x})|}.\vspace{-5pt}
\end{equation}
Fix $\alpha_0=a_3$, $\alpha_n=b_3$ and let $I=\{(\alpha_1,\dots,\alpha_{n-1})\,|\,a_3<\alpha_1<\dots<\alpha_{n-1}< b_3\}. $ Then the optimization problem is
\begin{align}
    &\min_{I} \text{KL}(P_x \| (h \circ B_n)_*P_z ) \\ =& - \max_{I} \int \log \frac{P_{z}(B_n^{-1}(h^{-1}\textbf{x}))}{|\textbf{J}_{h \circ B_n}(B_n^{-1}(h^{-1}\textbf{x}))|} \,\cdot\, P_x(\textbf{x})\, d\textbf{x} 
    \\ =&  - \max_{I} \int \log \frac{P_{z}(B_n^{-1}(h^{-1}\textbf{x}))}{|\textbf{J}_{h }(h^{-1}\textbf{x})\textbf{J}_{B_n}(B_n^{-1}(h^{-1}\textbf{x}))|} \cdot\, P_x(\textbf{x})\, d\textbf{x} 
    \\ =& - \max_{I} \left\{\int \log \frac{P_{z}(B_n^{-1}(h^{-1}\textbf{x}))}{|\textbf{J}_{B_n}(B_n^{-1}(h^{-1}\textbf{x}))|} \cdot\, P_x(\textbf{x})\, d\textbf{x}  + \int \log |\textbf{J}_{h }(h^{-1}\textbf{x})|\cdot\, P_x(\textbf{x})\, d\textbf{x}. \right\} \label{middle} 
    \\ =& - \max_{I}\left\{ \int \log \frac{P_{z}(B_n^{-1}(h^{-1}\textbf{x}))}{|\textbf{J}_{B_n}(B_n^{-1}(h^{-1}\textbf{x}))|} \cdot\, \frac{P_y(h^{-1}\textbf{x})}{|\textbf{J}_h(h^{-1}\textbf{x})|}\, d\textbf{x} \right\} + \int \log |\textbf{J}_{h }(h^{-1}\textbf{x})|\cdot\, \, \frac{P_y(h^{-1}\textbf{x})}{|\textbf{J}_h(h^{-1}\textbf{x})|} d\textbf{x}.
    \\ =& - \max_{I} \left\{\int \log \frac{P_{z}(B_n^{-1}(\textbf{y}))}{|\textbf{J}_{B_n}(B_n^{-1}(\textbf{y}))|} \cdot\, P_y(\textbf{y})\, d\textbf{y}\right\} + \int \log |\textbf{J}_{h}(\textbf{y})|\cdot\, P_y(\textbf{y})\, d\textbf{x}. 
    \\ =& \min_{I} \text{KL}(P_y \| (B_n)_*P_z ) +  \int \log |\textbf{J}_{h}(\textbf{y})|\cdot\, P_y(\textbf{y})\, d\textbf{x} \label{final}\vspace{-5pt}
\end{align}

Note that the second integral in \eqref{middle} can be taken outside the $\max$ because it is independent of $B_n$, and hence, it becomes a constant that is irrelevant for the optimization. From \eqref{final}, it follows that the minimum of $\text{KL}(P_x \| (h \circ B_n)_*P_z )$ is achieved if and only if the minimum of $\text{KL}(P_y \| (B_n)_*P_z )$ is achieved. Hence, 
\begin{equation}
    \text{arg min}_{I}\, \text{KL}(P_x \| (h \circ B_n)_*P_z )= \text{arg min}_{I}\, \text{KL}(P_y \| (B_n)_*P_z )
\end{equation}
as required. It is easy to see that this argument remains unchanged when $B_n$ is replaced by $f$ and $I$ is replaced by $f \in \mathfrak{F}$.
\end{proof}

\begin{proof}[Proof of Theorem 2]
There is a probabilistic interpretation of Bernstein polynomials that makes the analysis easier. Let $Z^x_i$, $0 \leq i \leq n$ be i.i.d.~Bernoulli$(x)$ random variables.  Then
\begin{equation}\label{BernProb}
    B_n(x)=\mathbb{E}\left(f\left(\sum_{i=0}^nZ^x_i/n\right)\right).
\end{equation}
See, for example, Chapter 2 of \citet{BernOp}. 
We will use this definition in the proof.

Let $f:[0,1]\to \mathbb{R}$ be a strictly increasing continuous function such that $f(k/n)=\alpha_k$. 
  Let $s<t$ and let $Z^x_i,\, 0 \leq i \leq n$ and be a sequence of iid Bernoulli$(x)$ for $x=s, t$, defined on the same probability space such that $Z^s_i \leq Z^t_i $ via monotone coupling. That is, let $Z^s_i= {\bf 1}_{U \leq s}$ and $Z^t_i={\bf 1}_{U \leq t}$ where $U$ is a uniform random variables on $[0,1]$ and couple them as follows.
\begin{equation}
  \mathbb{P}((Z^s_i,Z^t_i)=(j,k))_{j,k\in \{0,1\}}=\begin{pmatrix}
1-t & t-s \\ 0 & s
\end{pmatrix} 
\end{equation}
and $\mathbb{P}(Z^s_i > Z^t_i)=\mathbb{P}(Z^s_i=1, Z^t_i=0)=0$. So, $Z^t_i \geq Z^s_i$ as required.

Then
\begin{equation}
    f\left(\sum_{i=0}^n Z^s_i/n\right)\leq f\left(\sum_{i=0}^nZ^t_i/n\right).
\end{equation}
Consequently,
\begin{equation}
\label{equ:mono_eq2}
\mathbb{E}\left(f\left(\sum_{i=0}^n Z^s_i/n\right)\right) \leq \mathbb{E}\left(f\left(\sum_{i=0}^n Z^t_i/n\right)\right).
\end{equation}
Due to \eqref{BernProb}, this is equivalent to $B_n(s) \leq B_n(t)$. 

If \eqref{equ:mono_eq2} is not strict, then 
$f(\sum_{i=0}^n Z^t_i/n)=f(\sum_{i=0}^n Z^s_i/n)$ almost surely, and therefore, $\sum_{i=0}^n Z^t_i = \sum_{i=0}^n Z^s_i $ almost surely. But this is impossible due to monotone coupling. Therefore, by contradiction, \eqref{equ:mono_eq2} is strict as required.
\end{proof}

\begin{proof}[Proof of Theorem 3] Recall
From \citet{Bernstein} that $B_n$s are uniformly dense in the space of continuous function on $[0,1]$ because $B_n(f) \to f$ uniformly. By rescaling, this is true on any interval $[a,b]$. Moreover, by construction, whenever $f$ is increasing, $B_n(f)$ is increasing. So, it is automatic that increasing Bernstein polynomials on $[a,b]$ are uniformly dense in the space of increasing continuous functions on $[a,b]$. Finally, to show true universality, we have to show that any increasing 
continuous function $f:\mathbb{R} \to \mathbb{R}$ is well-approximated by $B_n$s. 

Given $f:\mathbb{R} \to \mathbb{R}$ continuous and increasing, choose two positive sequences $\{M_n\}$ and $\{\varepsilon_n\}$ such that $M_n \to \infty$ and $\varepsilon_n \to 0$. Let $I_n=[-M_n, M_n]$. Then, there exists a Bernstein approximation of $f$, say $q_n$, which is increasing on $I_n$ (which can be monotonically extended to $\mathbb{R}$) such that \vspace{-5pt}
\begin{equation}
    \max_{I_n} |f-q_n| \leq \varepsilon_n.\vspace{-5pt}
\end{equation}
Then the sequence of Bernstein approximations $\{q_n\}$ converges point-wise to $f$ on $\mathbb{R}$, and this convergence is uniform on each compact interval. 
\end{proof}

\begin{remark}\label{rem:MonoApprox}
We can write down a sequence $q_n$ explicitly when $f$ is regular. For example, when $f$ is $C^3$ with bounded derivatives and $M_n=\log n$, choosing the degree of $q_n$ to be $n$ is sufficient because it follows from the error estimate in Section 4.3 that $\varepsilon_n \sim (\log n)/n$ works. That is, choose $q_n$ to be the degree $n$ Bernstein approximation of $f$ on $[-M_n,M_n]$. 
\end{remark}

\begin{remark}
Note that this result is not a restatement of the original result in \cite{Bernstein}. The latter is about Bernstein-type polynomials being uniformly dense in the space of continuous functions on a \textit{compact} interval. It uses the fact that such functions have a maximum. For the universality of NFs, we need that given an increasing continuous function on the real line (which is noncompact and hence, no guarantee of a maximum) there is a sequence of Bernstein-type polynomials that converge (at least, pointwise) to it. 
\end{remark}\vspace{-15pt}

\section{Universality and the explicit rate of convergence}\vspace{-5pt}
The basis of all the universality proofs of NFs in the existing literature is that the learnable class of functions is dense in the class of increasing continuous functions. In contrast, the argument we present here is constructive. As a result, we can write down sequences of approximations for (known) transformations between densities; see \Cref{sec:examples}.

In the case of cubic-spline NFs of \citet{durkan2019cubic}, it is known that for $k=1,2,3$ and $4$, when the transformation is $k$ times continuously differentiable and the bin size is $h$, the error is $O(h^k)$ \citep[Chapter 2]{SplinesBook}. However, we are not aware of any other instance where an error bound is available. 
Fortunately for us, the error of approximation of a function $f$ by its Bernstein polynomials has been extensively studied. 
We recall from \citet{Voronovskaya} the following error bound: for $f:[0,1] \to \mathbb{R}$ twice continuously differentiable
\begin{equation}
    B_n(f)-f=\frac{x(1-x)}{2n}f''(x)+o(n^{-1}).
\end{equation}
and this holds for an arbitrary interval $[a,b]$ with $x(1-x)$ replaced by $(x-a)(b-x)$.


Since the error estimate is given in terms of the degree of the polynomials used,
we can improve the optimality of our NF by avoiding unnecessarily high degree polynomials. This allows us to keep the number of trainable parameters under control in our NF model.
The following example shows that the error $O(n^{-1})$ 
above does not necessarily improve when SOS polynomials are used instead. 
\begin{exmp}\label{eg4}
\emph{Uniform}$[0,1]$ to the \emph{Normal}$(0,1):$\,There is bounded $\{c_k\}_{k\geq 0} \subset\mathbb{R}_+$ such that
\begin{align}
    f(z)=\emph{Erf}^{-1}(2z-1) = \sum_{k=0}^\infty \frac{\sqrt{2}\pi^{k+\frac{1}{2}} c_k}{2k+1}\left(z-\frac{1}{2}\right)^{2k+1};
\end{align}
see \emph{\citet{jaini2019sum}}. This is the power series expansion of $f$ at $z=1/2$, and hence, it is unique. The SOS approximation 
of $f$ $($the series above truncated at $k=n)$ is only $O((2n+1)^{-1})=O(n^{-1})$ accurate on compact sub-intervals of $(0,1)$. This is precisely the accuracy one would expect from the degree $2n+1$ Bernstein approximation on any compact subinterval of $(0,1)$.
\end{exmp}
In our NF, at each step, the estimation is done using a univariate polynomial, and hence, the overall convergence rate is, in fact, the minimal univariate convergence rate of $O(n^{-1})$ (equivalently, the error upper bound is the maximum of univariate upper bounds),
and in general, cannot be improved further regardless of how regular the density transformation is. However, our experiments 
show that our model on average has a significantly smaller error than the given theoretical upper-bound. \vspace{-10pt}

\section{Examples of Bernstein-type approximations}\label{sec:examples}\vspace{-5pt}
In this section, we illustrate how to use Bernstein-type polynomials to approximate diffeomorphisms between densities. We restrict our attention to densities on $\mathbb{R}$. 
Suppose $F$ and $G$ are the distribution functions of the two probability densities $P_z$ and $P_x$ on $\mathbb{R}$. Then the \textit{increasing rearrangement} $f=G^{-1}\circ F$ is the unique increasing transformation that pushes forward $P_z$ to $P_x$, and this 
generalizes to higher dimensions \citep[Chapter 1]{Villani}. Now, we can explicitly write down their degree$-n$ Bernstein-type approximations, $B_n(f)$ along with convergence rates.
\begin{exmp}
\emph{Uniform}$[0,1]$ to a continuous and non-zero density $P$ on $[0,1]:$ Note that  $G(x)=\int_0^x P(s) \, ds,\,x \in [0,1]$ is strictly increasing and hence, invertible on $[0,1]$. So, $f(x)=G^{-1}(x)$, and $G^{-1}$ is once continuously differentiable. Then\vspace{-5pt} 
\begin{equation}
    B_n(f)(x)= \sum_{k=0}^nG^{-1}\left(\frac{k}{n}\right)\binom{n}{k}x^k(1-x)^{n-k}.\vspace{-5pt}
\end{equation}
and $\|B_n(f)- f\|_{\infty}= O(n^{-1/2})$.   
\end{exmp}
\begin{exmp}\label{eg:Kumar}
\emph{Kumaraswamy}$(\alpha,\beta)$ to \emph{Uniform}$[0,1]:$ Here, $\alpha, \beta >0$ and for $x\in [0,1]$, $F(x)=1-(1-x^\alpha)^\beta$ \emph{~\citep{Kumar}} and $G(x)=x$. Therefore, $f(x)=F(x)$. Then \vspace{-5pt} 
\begin{equation}
    B_n(f)(x)= \sum_{k=0}^n F\left(\frac{k}{n}\right)\binom{n}{k}x^k(1-x)^{n-k}.\vspace{-5pt}
\end{equation}
When $\alpha, \beta \geq 1$, $\|B_n(f)-f\|=O(n^{-1})$. 
\end{exmp}

\section{Hyper-parameters and training details}\vspace{-5pt}

For optimization, we used the Adam optimizer with parameters $\beta_1 = 0.9$, $\beta_1 = 0.999$, $\varepsilon = 1 \times 10^{-8}$, where parameters refer to the usual notation. An initial learning rate of $0.01$ was used for updating the weights with a decay factor of $10\%$ per $50$ iterations. We initialized all the trainable weights randomly from a standard normal distribution and used maximum likelihood as the objective function for training. We observed that a single layer model with $100$ degree polynomials performed well for the real-world data. 

In contrast, for 2D toy distributions and and images we used higher number of layers ($8$) with $15$ degree polynomials in each layer. For all the experiments, we use a Kumaraswamy distribution with parameters $a = 2$ and $b = 5$ as the base density. 
Using a standard normal distribution after converting it to a density on $[0,1]$ using a nonlinear transformation, e.g., $\frac{1 + \mathrm{tanh}(z)}{2}$, also yielded similar results. \vspace{-10pt}

\section{Training stability for higher degree polynomials}\label{sec:degree}\vspace{-5pt}

\begin{figure}[!ht]%
\centering 
\subfigure[\textsc Power]{
\includegraphics[width=0.45\linewidth]{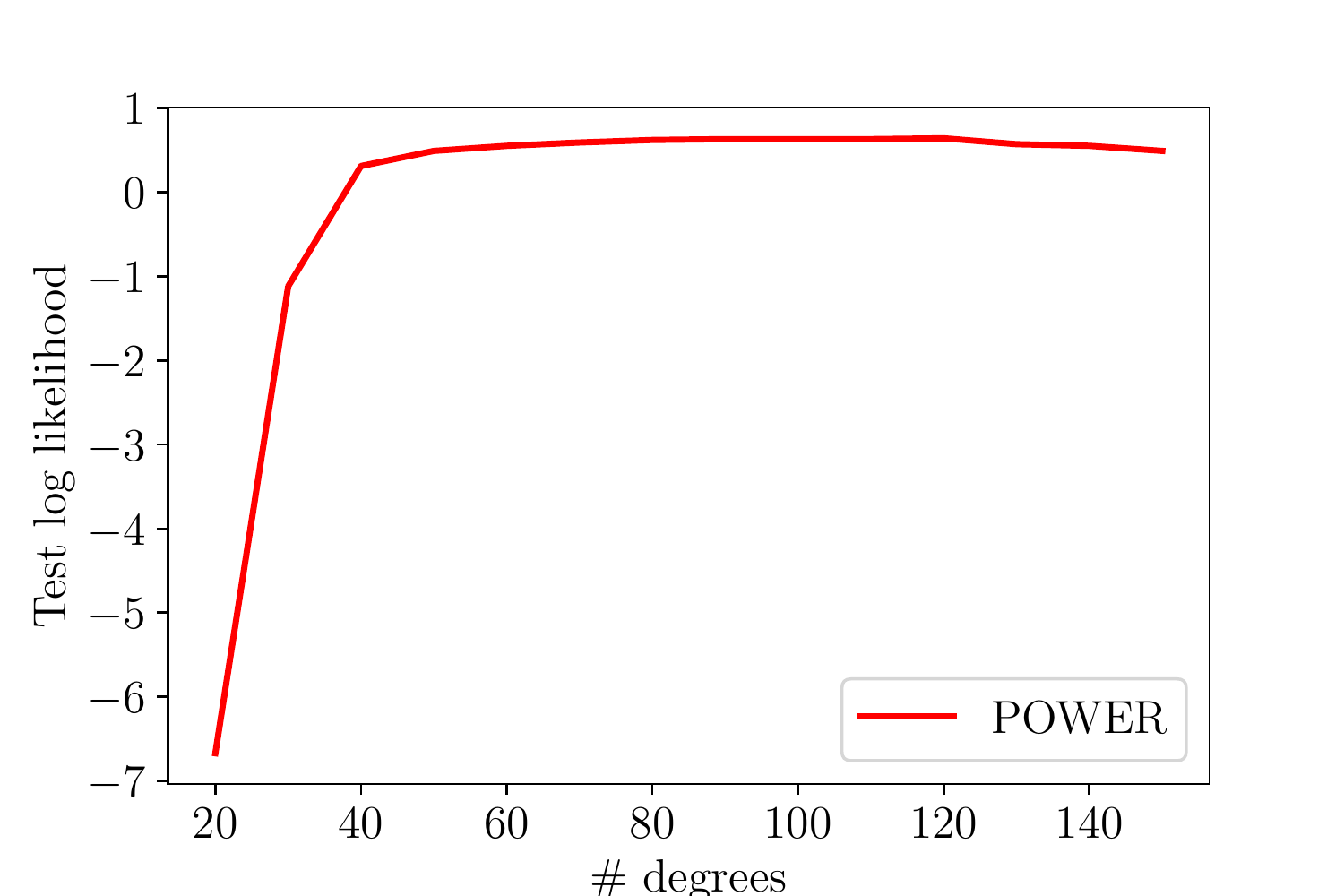} }
\subfigure[\textsc Gas]{
\includegraphics[width=0.45\linewidth]{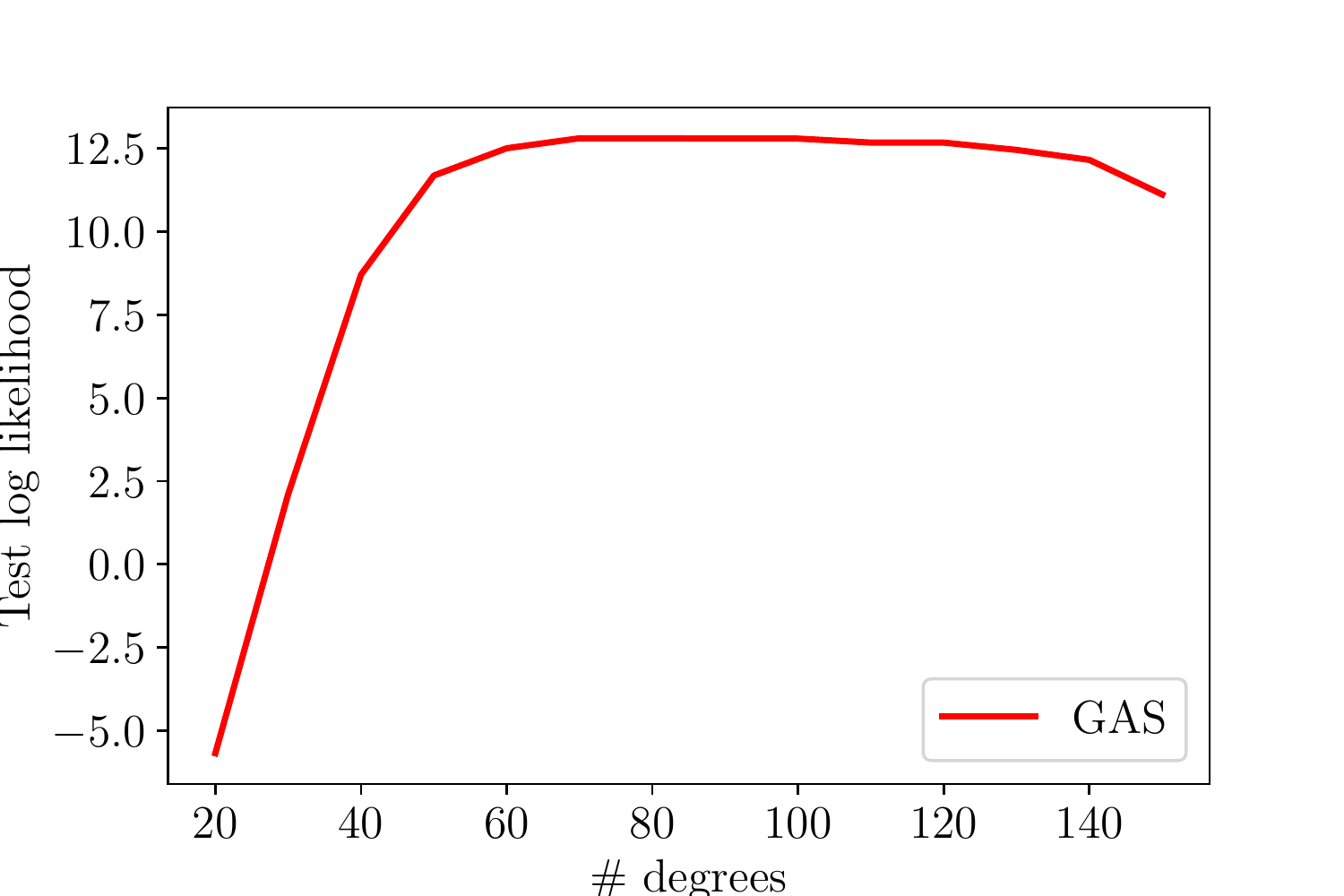} }
\caption{Test log-likelihood against the number of degrees used for the Bernstein approximation in a single layer model on \textsc{Power} and \textsc{Gas} datasets. Slight dip in the performance for degrees 100+ but shows no training instability. }%
\label{fig:degree}\vspace{-5pt}
\end{figure}

Typically, polynomial-based models such as SOS yield training instability as their target ranges are not compact. 
This is because higher degree approximations could increase the range of outputs without bound, and in turn cause gradients to explode while training. As a solution, they opt to use a higher number of layers with lower degree polynomials. In contrast, our model can entertain higher degree approximations without any instability which allows more design choices. \Cref{fig:degree} demonstrates this behavior experimentally. \vspace{-15pt}

According to \Cref{fig:degree}, the model hits a peak in performance at a certain degree and shows a slight drop in performance at higher degrees. Nevertheless, the model does not exhibit unstable behavior at higher degrees as opposed to SOS-flows -- an indication of the superior training stability of our model. This further illustrates that our model provides the option to design shallow models by increasing the number of degrees in the polynomials instead of deeper models with a higher number of layers.\vspace{-10pt}

\section{Ablation study}\vspace{-5pt}

\begin{figure*}[!ht]
    \centering
    \includegraphics[width=1\columnwidth]{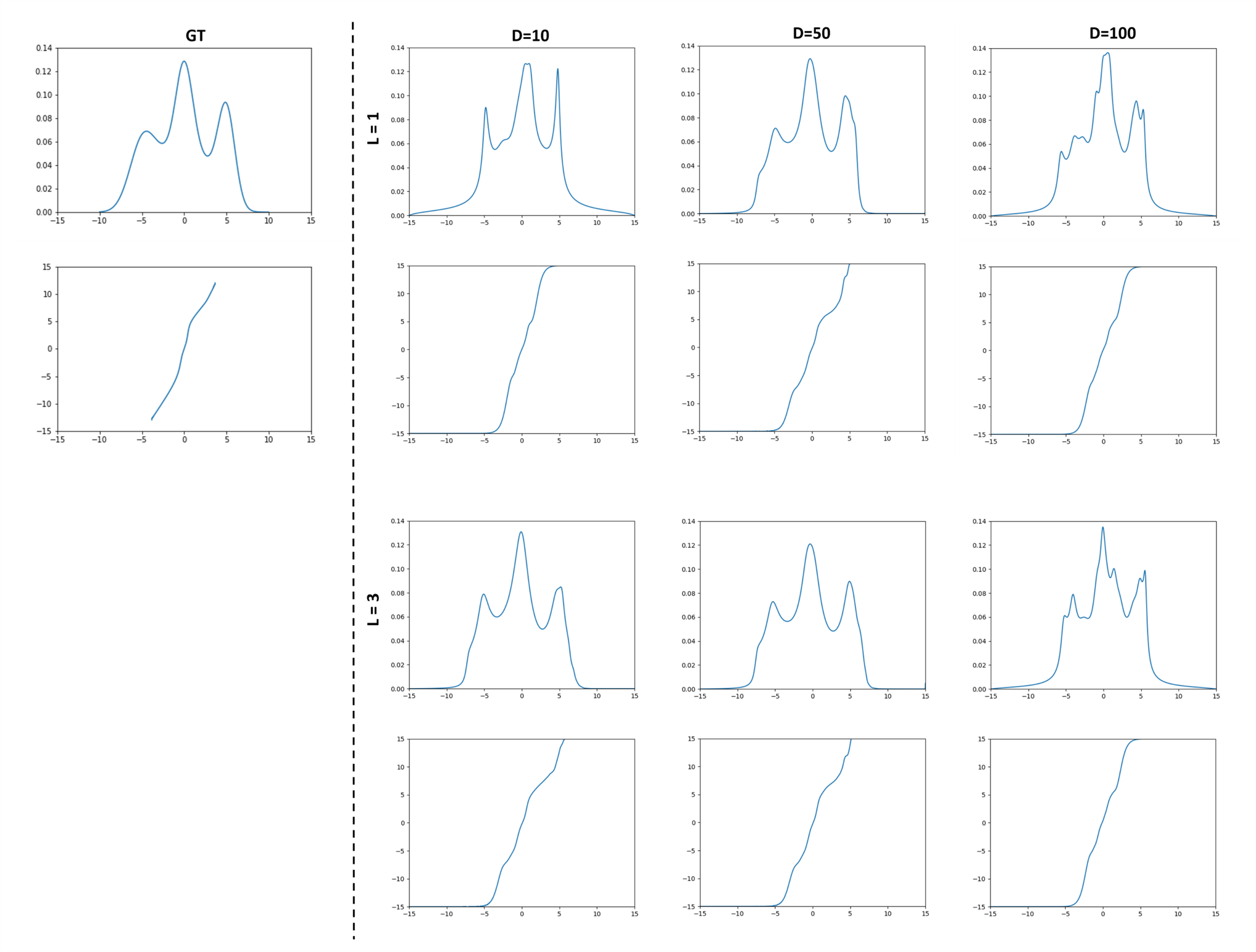}
    \vskip -0.1in
    \caption{\small Ablation study with different varients of our model. \textbf{D} and \textbf{L} denotes the degree of the used polynomials and the number of layers, respectively. Corresponding transformation functions are also shown below the predicted densities.}\vspace{-10pt}
    \label{fig:ablation_study}
\end{figure*}

We compare the performance of different variants of our model against a simple task in order to better understand the design choices. For this, we use a standard normal as the base distribution, and a mixture of five Gaussians with means $= (-5,-2,0,2,5)$, variances = $(1.5,2,1,2,1)$, and weights $0.2$ each, as the target. \Cref{fig:ablation_study} depicts the results.

Clearly, we were able to increase the expressiveness of the transformation by increasing the degree of the polynomials, as well as the number of layers. However, it is also visible that using an unnecessarily higher degree over-parametrizes the model, and hence, deteriorate the output. 
As discussed in the main article and in \Cref{sec:degree}, we are able to use polynomials with degree as high as 100 in this experiment and others with no cost to the training stability because the training is done for a compactly supported target. 

\begin{figure*}[!ht]
    \centering
    \includegraphics[width=1\columnwidth]{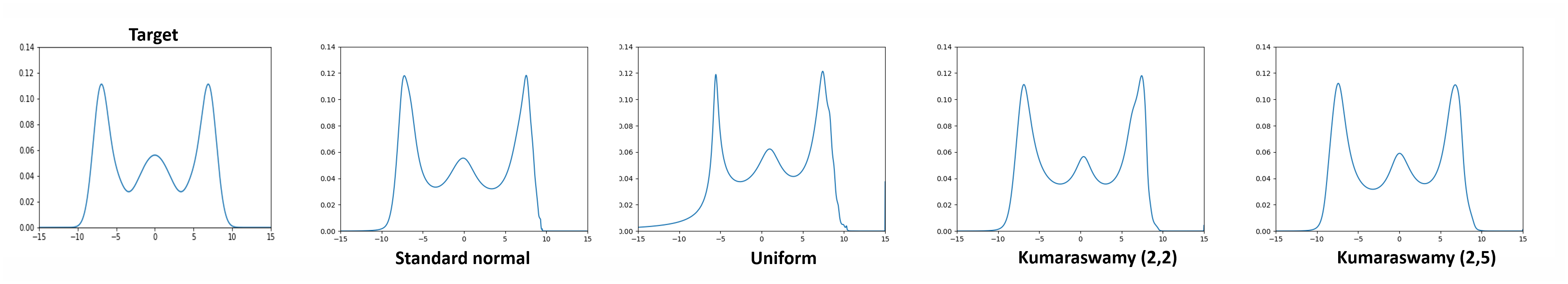}
    \vskip -0.1in
    \caption{\small Approximation of the target density starting from various initial densities (the initial distributions are noted below the densities).}\vspace{-5pt}
    \label{fig:ablation_base}
\end{figure*}

We also examine how the initial base distribution affects the performance. We use a mixture of seven Gaussians with means $= (-7, -5,-2,0,2,5,7)$, variances = $(1,1,2,2,2,1,1)$, and weights $= (0.8, 0.2, 0.2, 0.6, 0.2, 0.2, 0.8)$, as the target. We used a model with a $100$-degree polynomial and a single layer for this experiment. \Cref{fig:ablation_base} illustrates the results.  
Although all priors capture the multimodes, when Uniform$[0,1]$ is used the model was not able to predict that the density is almost zero for large negative values. \vspace{-10pt} 


%




\vspace{-10pt}

\section{Experiments with two-dimensional image datasets}

\begin{figure}[!h]%
\centering 
\subfigure[\textsc MNIST]{
\includegraphics[width=0.4\linewidth]{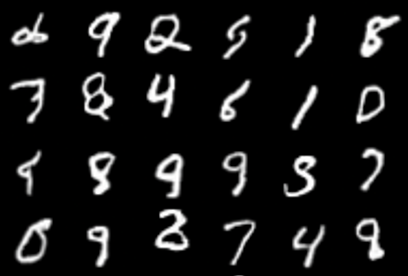} }
\subfigure[\textsc CIFAR10]{
\includegraphics[width=0.4\linewidth]{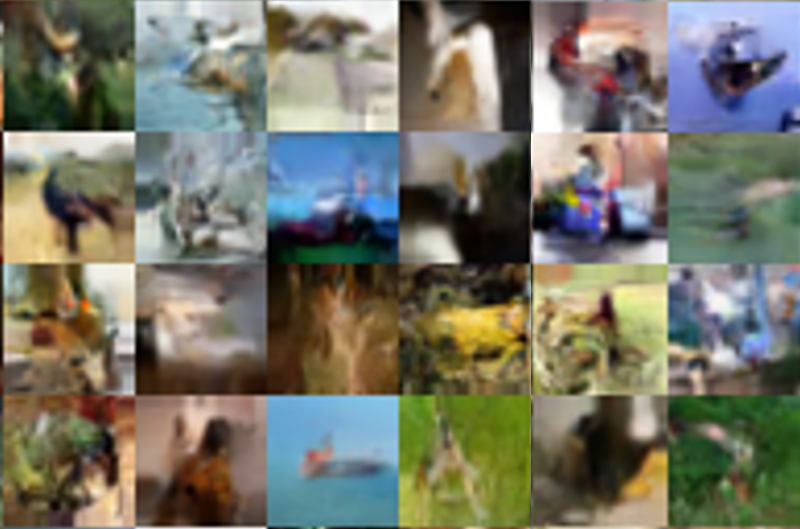} }
\caption{Generated samples on MNIST and CIFAR10.}%
\label{fig:images}
\end{figure}

In Section 4.2, we reported the results of the experiments we ran on two low-dimensional image datasets:~CIFAR10 and MNIST. The samples generated from the experiment are shown \Cref{fig:images}. We recall that we manage to obtain the optimal results among the methods that do not use multi-scale convolutional architectures. 

\bibliography{main}

\end{document}